\documentclass{article} 
\usepackage{nips13submit_e,times}
\usepackage{url}
\usepackage{graphicx} 
\usepackage{subfigure} 
\usepackage{amssymb}
\usepackage{amsmath}
\usepackage{verbatim}
\usepackage{multirow}
\usepackage{tikz}

\usepackage{algorithm}
\usepackage{algorithmic}
\newtheorem{example}{Example}[section]

\newcommand{\specialcell}[2][c]{%
  \begin{tabular}[#1]{@{}c@{}}#2\end{tabular}
}
\title{Adaptive Feature Ranking for Unsupervised Transfer Learning}

\author{
Son N. Tran \\
Department of Computer Science\\
City University London\\
London, UK, EC1V 0HB \\
\texttt{Son.Tran.1@city.ac.uk} \\
\And
Artur d'Avila Garcez \\
Department of Computer Science \\
City University London \\
London, UK, EC1V 0HB\\
\texttt{aag@soi.city.ac.uk} \\
}

\nipsfinalcopy 

\begin{document}

\maketitle

\begin{abstract} 

Transfer Learning is concerned with the application of knowledge gained 
from solving a problem to a different but related problem domain. 
In this paper, we propose a method and efficient algorithm for ranking and 
selecting representations from a Restricted Boltzmann Machine trained on 
a source domain to be transferred onto a target domain. 
Experiments carried out using the MNIST, ICDAR and TiCC image datasets show
that the proposed adaptive feature ranking and transfer learning method offers
statistically significant improvements on the training of RBMs.
Our method is general in that the knowledge chosen by the ranking function does 
not depend on its relation to any specific target domain, and it works with 
unsupervised learning and knowledge-based transfer.

\end{abstract} 

\section{Introduction}

Transfer Learning is concerned with the application of knowledge
gained from solving a problem to a different but related problem
domain.  A number of researchers in Machine Learning have argued that
the provision of supplementary knowledge should help improve learning
performance \cite{Towel_1994, Garcez_1999, Richardson_2006,
  Argyriou_2007, Raina_2007, Torrey_2010, Mihalkova_2007, Davis_2009}.
In Transfer Learning \cite{Pan_2010,Argyriou_2007, Raina_2007,
  Torrey_2010}, knowledge from a source domain can be used to improve
performance in a target domain by assuming that related domains have
some knowledge in common.  In connectionist transfer learning, many
approaches transfer the knowledge selected specifically based on the
target, and (in some cases) with the provision of labels in the source
domain \cite{Mesnil_2012, WeiP_2011}. In constrast, we are interested in selecting the
representations that can be transferred in general to the target
without the provision of labels in the source domain, which is similar
to self-taught learning \cite{Raina_2007,Lee_2008}. In addition, we
propose to study how much knowledge should be transferred to the
target domain, which has not been studied yet in self-taught mode.

This paper introduces a method and efficient algorithm for ranking and 
selecting representation knowledge from a Restricted Boltzmann Machine (RBM) 
\cite{Smolensky86, Hinton_2002} trained on a source domain to be
transferred onto a target domain. A ranking function is defined that is 
shown to minimize information loss in the source RBM. High-ranking features 
are then transferred onto a target RBM by setting some of the network's parameters.
The target RBM is then trained to adapt a set of additional parameters on a 
target dataset, while keeping the parameters set by transfer learning fixed. 



Experiments carried out using the MNIST, ICDAR and TiCC image datasets show
that the proposed adaptive feature ranking and transfer learning method offers
statistically significant improvements on the training of RBMs in three out of four 
cases tested, where the MNIST dataset is used as source and either the ICDAR or 
TiCC dataset is used as target. As expected, our method improves on 
the predictive accuracy of the standard self-taught learning method 
for unsupervised transfer learning \cite{Raina_2007}. 

Our method is general in that the knowledge chosen by the ranking
function does not depend on its relation to any specific target
domain. In transfer learning, it is normal for the choice of the
knowledge to be transferred from the source domain to rely on the
nature of the target domain. For example, in \cite{Lim_2011},
knowledge such as data samples in the source domain are transformed to
a target domain to enrich supervised learning.  In our approach, the
transfer learning is unsupervised, as done in \cite{Raina_2007}.  We
are concerned with inductive transfer learning in that representation
learned from a domain can be useful in analogous domains. For example,
in \cite{Argyriou_2007} a common orthonormal matrix is learned to
construct linear features among multiple tasks. Self-taught learning
\cite{Raina_2007}, on the other hand, applies sparse coding to
transfer the representations learned from source data onto the target
domain. Like self-taught, we are interested in unsupervised transfer
learning using cross-domain features. The system has been implemented
in MATLAB; all learning parameters and data folds are available upon
request.

The outline of the paper is as follows. In Section 2, we define feature selection by ranking in which high scoring features are associated with significant part of the network.
In Section 3, we introduce the adaptive feature learning method and algorithm to combine selective features in source domain with features in target domain.
In Section 4, we present and discuss the experimental results. Section 5
concludes and discusses directions for future work.

\section{Feature Selection by Ranking}
\label{sec:ranking}
In this section, we present a method for selecting representations from an RBM 
by ranking their features. We define a ranking function and show that it can 
capture the most significant information in the RBM, according to a measure
of information loss.

In a trained RBM, we define $c_j$ to be a score for each unit $j$ in the hidden
layer. The score represents the weight-strength of a sub-network $N_j$ ($h_j
\sim V$) consisting of all the connections from unit $j$ to the visible layer. 
A score $c_j$ is a positive real number that can be seen as capturing the uncertainty in 
the sub-network. We expect to be able to replace all the weights of a network by
appropriate $c_j$ or $-c_j$ with minimum information loss. We define
the information loss as:
\begin{equation}
\label{eq:op_f}
\mathcal{I}_{loss} = \sum_j\|\mathbf{w}_j - c_j\mathbf{s}_j\|^2
\end{equation}
\noindent where $\mathbf{w}_j$ is a vector of connection weights from unit $j$
in the hidden layer to all the units in the visible layer, $\mathbf{s}_j$ is a
vector of the same size $visN$ as $\mathbf{w}_j$, and $\mathbf{s}_j
\in [-1 \quad 1]^{visN}$.

Since equation \eqref{eq:op_f} is a quadratic function, the value of $c_j$ that 
minimizes information loss can be found by setting the derivatives to zero, as follows:

\begin{equation}
\begin{aligned}
&\sum_i 2(w_{ij} - c_js_{ij})s_{ij} = 0 \text{ with all } j \\
& \sum_i w_{ij}s_{ij} - c_j\sum_i s^2_{ij} = 0 \text{ with all } j \\
& c_j = \frac{\sum_i w_{ij}s_{ij}}{\sum_i s^2_{ij}} \\
\label{eq:cfd_1}
\end{aligned}
\end{equation}

Since $s_{ij} = [-1 \quad 1]$ and from equation \eqref{eq:op_f}, we can see that:

\begin{equation}
\|w_{ij} - c_js_{ij}\|^2 = \|abs(w_{ij}) - c_j\frac{s_{ij}}{sign(w_{ij})}\|^2 \ge \|abs(w_{ij}) - c_j\|^2
\label{eq:cfd_3}
\end{equation}

holds if and only if $s_{ij} = sign(w_{ij})$, which will also minimize $\mathcal{I}_{loss}$. 

Applying equation \eqref{eq:cfd_3} to \eqref{eq:cfd_1}, we obtain:

\begin{equation}
c_j = \frac{\sum_i abs(w_{ij})}{visN}
\end{equation}

We may notice that using $c_j$ instead of weights would result in a
compression of the network; however, in this paper we focus on the
scores only for selecting features to use for transfer learning. In what
follows, we give a practical example which shows that low-score features
are semantically less meaningful than high-scoring ones.

\begin{example}
  We have trained an RBM with 10 hidden nodes to model the XOR function from its
  truth-table such that $z= x \text{ XOR } y$. The $true,false$ logical values are represented by
  integers $1,0$, respectively. After training the network,
  a score for each sub-network can be obtained. The score allows us to replace each
  real-value weight by its sign, and interpret those signs logically where a negative sign 
  ($-$) represents logical negation ($\neg$), as exemplified in Table \ref{tab:xor_interpret}. 

\begin{table}[ht]
\centering
\begin{tabular} {c l c}
Network & Sub-network & Symbolic representation\\
\hline
\includegraphics[width = 0.25\textwidth]{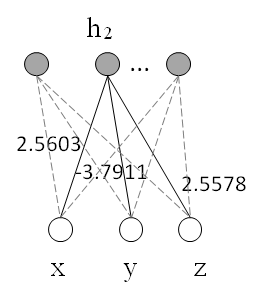} & 
\specialcell[b]{\includegraphics[width = 0.18\textwidth]{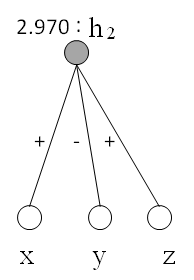}}
&\specialcell[b]{ $2.970: h_2 \sim \{+x,-y,+z\}$ \\\\ can be interpreted as \\\\ $z = x \wedge \neg y $ \\\\ with score $2.970$  \\\\ }\\
\hline
\end{tabular}
\caption{RBM trained on XOR function and one of its sub-networks with score value and logical interpretation}
\label{tab:xor_interpret}
\end{table}  

Table \ref{tab:xor-extract} shows all the sub-networks with 
associated scores. As one may recognize, each sub-network 
represents a logical rule learned from the RBM.
However, not all of the rules are correct w.r.t. the XOR function. 
In particular, the sub-network scoring $0.158$ encodes a rule 
$\neg z = \neg x \wedge y$, which is inconsistent with  $z= x \text{ XOR } y$. 
By ranking the sub-networks according to their scores, this can be identified: 
high-scoring sub-networks are consistent with the data, and low-scoring ones are not.
We have repeated the training several times with different numbers of hidden
units, obtaining similar intuitive results.

\begin{table}[h]
\centering
\begin{tabular} {|c|l|c|}
\hline\hline
Score & Sub-network & Logical Representation\\
\hline
$1.340$ & $h_1 \sim \{+x,-y,+z\}$   & $z = x \wedge \neg y$ \\
$2.970$ & $h_2 \sim \{+x,-y,+z\}$   & $z = x \wedge \neg y$\\
$6.165$ & $h_3 \sim \{-x,+y,+z \}$  & $z = \neg x \wedge y$\\
{\color{red}$0.158$}& {\color{red}$h_4 \sim \{-x,y,-z \}$}& {\color{red}$\neg z = \neg x \wedge y$}\\
$2.481$ & $h_5 \sim \{+x,-y,+z\}$   & $z = x \wedge \neg y$	\\
$1.677$ & $h_6 \sim \{-x,+y,+z\} $  & $z = \neg x \wedge y$ \\
$2.544$ & $h_7 \sim \{+x,-y,+z\}$   & $z = x \wedge \neg y$\\
$7.355$ & $h_8 \sim \{+x,+y,-z\}$   & $\neg z = x \wedge y $\\
$6.540$ & $h_9 \sim \{-x,-y,-z\}$   & $\neg z = \neg x \wedge \neg y$\\
$4.868$ & $h_{10}\sim \{-x,-y,-z$   & $\neg z = \neg x \wedge \neg y$\\
\hline
\end{tabular}
\caption{Sub-networks and scores from RBM with 10 hidden 
units trained on XOR truth-table}
\label{tab:xor-extract}
\end{table}  

\end{example}

In what follows, we will study the effect of pruning low-scoring sub-networks 
from an RBM trained on complex image domain data. So far, we have seen that 
our scoring can be useful at identifying relevant representations. 
In particular, if the score of a sub-network is small in relation to the average 
score of the network, they seem more likely to represent noise.

\section{Adaptive Feature Learning}
\label{subsec:learning}
\begin{figure}[ht]
\centering
\includegraphics[width=0.6\textwidth]{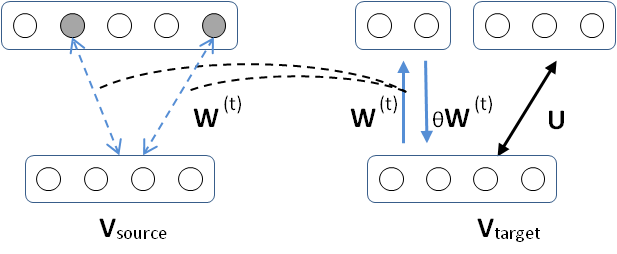}
\caption{General adaptive feature transfer mechanism for unsupervised learning}
\label{fig:model}
\end{figure}

In this section, we introduce adaptive feature learning. 
Given an RBM trained on a dataset, we are interested in investigating 
whether the score ranking introduced in Section \ref{sec:ranking} can be useful 
at improving the predictive accuracy of another RBM trained on a target domain. 
From a transfer learning perspective, we intend to produce a general transfer 
learning algorithm, ranking knowledge from a source domain to guide the 
learning on a target domain, using RBMs as transfer medium. 
In particular, we propose a transfer
learning model as shown in Figure \ref{fig:model}. Knowledge learned
from a source domain is selected by the ranking function for transferring onto 
a target domain, as explained in what follows. The selection of features in the 
source domain is general in that it is independent from the data from the target 
domain.

In the target domain, an RBM is trained given a number of transferred
parameters ($W^{(t)}$): a fixed set of weights associated with
high-ranking sub-networks from the source domain. The output of
transferred hidden nodes in target domain is considered as self-taught
features \cite{Raina_2007}. The connections between the visible layer
and the hidden units transferred onto the target RBM can be seen as a
set of up-weights and down-weights. How much the down-weights affect
the learning in the target RBM depends on an influence factor
$\theta$; in this paper $\theta \in [0 \quad 1]$. If $\theta=0$ then
the features are combined but the transferred knowledge will not
influence learning in the target domain. In this case, the result is a
combination of self-taught features and target RBM
features. Otherwise, if $\theta=1$ then the transferred knowledge will
influence learning in the target domain. We refer to this case as
adaptive feature learning, because the knowledge transferred from the
source domain is used to guide the learning of new features in the
target domain. The outputs of additional hidden nodes (associated with
parameter $U$) in target RBM are called adaptive features.

As usual, we train the target RBM to maximize the log-likelihood:
\begin{equation}
\mathcal{L}(U|\mathcal{D};W^{(t)})
\end{equation}
using Contrastive Divergence \cite{Hinton_2002}.
\begin{algorithm}[ht]
  \caption{Adaptive feature learning
    \label{alg:transfer-learning}}
  \begin{algorithmic}[1]
    \REQUIRE{A trained RBM $N$}
    \STATE Select a number of sub-networks  $N^{(t)} \in N$ with the highest scores
    \STATE Encode parameters $W^{(t)}$ from $N^{(t)}$ into a new RBM
    \STATE Add hidden units ($U$) to the RBM
    \LOOP 
    \STATE{\%  until convergence}
{\footnotesize
      \STATE  $V_{+} \Leftarrow X$
      \STATE
         $H_{+} \gets p(H|V_{+}); \quad \quad \hat{H}_{+} \stackrel{smp}{\gets} p(H|V_{+});$ 
      \STATE 
         $V_{-} \gets p(V|\hat{H}_{+}); \quad \quad \hat{V}_{-} \stackrel{smp}{\gets} p(V|\hat{H}_{+})$
      \STATE
         $H_{-} \gets p(H|\hat{V}_{-})$
      \STATE
         $Y = U + \eta( \langle V^T_{+} H^{(t)}_{+} -  \hat{V}^T_{-} H^{(t)}_{-} \rangle)$
}
    \ENDLOOP
\end{algorithmic}
\end{algorithm}
\section{Experimental Results}
\label{sec:experimental_result}
We start by evaluating the approach on the MNIST handwritten dataset.
First, we want to check if hidden units with low scores are indeed
less significant in the case of image domains. Subsequently, we show
that knowledge from one image domain can be used to improve predictive
accuracy in another image domain.

\subsection{Feature Selection}

\begin{figure}[ht]
\centering
\subfigure[Filter bases with high scores] {
     \includegraphics[width=0.45\textwidth]{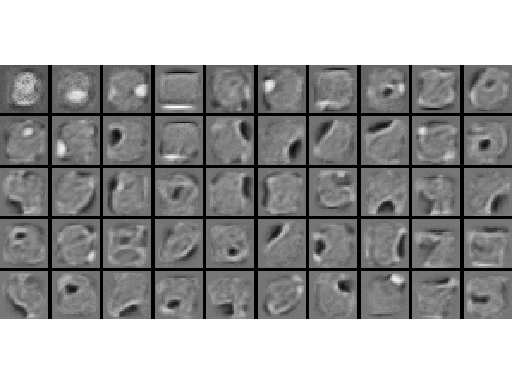}  
    }
\subfigure[Filter bases with low scores] {
     \includegraphics[width=.45\textwidth]{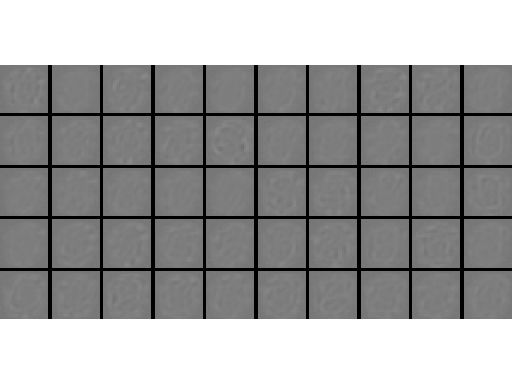}  
    }
\caption{Features learned from RBM on MNIST dataset}
\label{fig:ranking_rbm}
\end{figure}

\begin{figure}[ht]
\centering
\subfigure[Filter bases with high scores]{
     \includegraphics[width=0.45\textwidth]{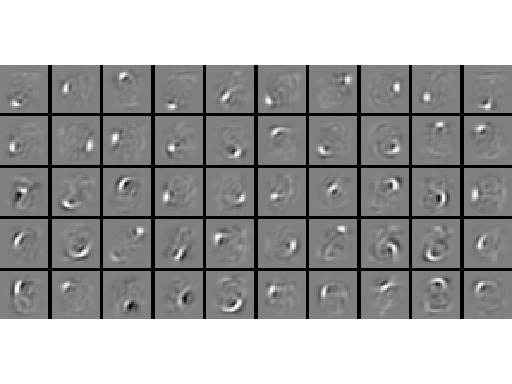}  
    }
\subfigure[Filter bases with low scores] {
     \includegraphics[width=.45\textwidth]{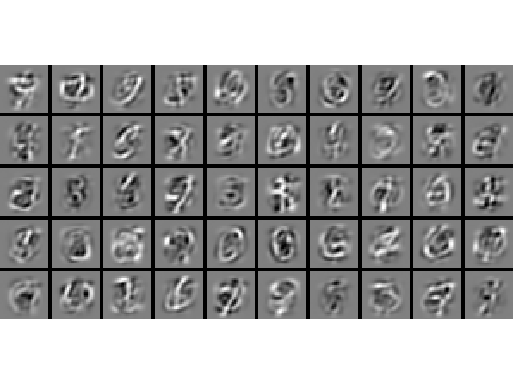}  
    }
\caption{Learned features from sparse RBM  on MNIST dataset}
\label{fig:ranking_srbm}
\end{figure}

We have trained an RBM with 500 hidden nodes on 20,000 samples from
the MNIST dataset in order to visualized the filter bases of the $50$
highest scoring sub-networks and the $50$ lowest scoring sub-networks
(each takes $10\%$ of the network's capacity). Figure
\ref{fig:ranking_rbm} shows the result of using a standard RBM, and
Figure \ref{fig:ranking_srbm} shows the result of using a sparse RBM
\cite{Lee_2008}.  As can be seen, in Figure \ref{fig:ranking_rbm}, high scores
are mostly associated with more concrete visualizations of the
expected MNIST patterns, while low scores are mostly associated with
fading or noisy patterns. In Figure \ref{fig:ranking_srbm},
high-scores are associated with sparse representations, while
low-scores produce less meaningful representations according to the
way in which the RBM was trained. In sparse RBM we use PCA to reduce
the dimensionality of the images to $69$ and train the network with
Gaussian visible units.

\begin{figure}[ht]
  \centering
    \includegraphics[width=0.15\textwidth]{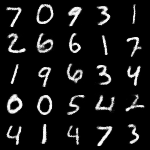}
    \hskip 0.1cm
    \includegraphics[width=0.15\textwidth]{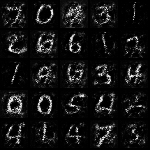}
    \hskip 0.1cm
    \includegraphics[width=0.15\textwidth]{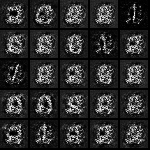}
    \hskip 0.1cm
    \includegraphics[width=0.15\textwidth]{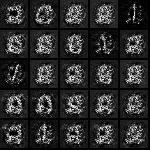}        
    \hskip 0.1cm
    \includegraphics[width=0.15\textwidth]{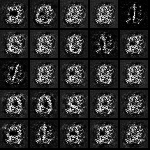}        
    \caption{Reconstructed test images from RBM in which high-scored
      features have been pruned.  From left to right, number of hidden
      unit remain: 500 (full), 400 ,300, 200 and 100}
\label{fig:inv_prune_vis}
\end{figure}

\begin{figure}[ht]
  \centering
    \includegraphics[width=0.15\textwidth]{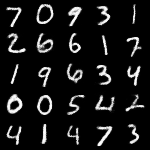}
    \hskip 0.1cm
    \includegraphics[width=0.15\textwidth]{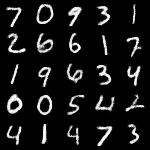}
    \hskip 0.1cm
    \includegraphics[width=0.15\textwidth]{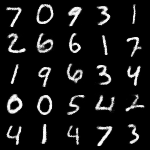}
    \hskip 0.1cm
    \includegraphics[width=0.15\textwidth]{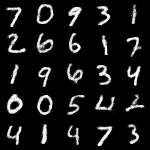}        
    \hskip 0.1cm
    \includegraphics[width=0.15\textwidth]{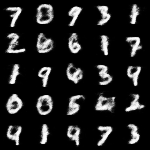}        
    \caption{Reconstructed test images from RBM in which low-scored
      features have been pruned.  From left to right, number of hidden
      unit remain: 500 (full), 400 ,300, 200 and 100}
\label{fig:prune_vis}
\end{figure}

We also examined visually the \emph{impact} of the scores on an RBM
with 500 hidden units trained on the MNIST dataset's 10,000
examples. Sub-networks with the highest scores were gradually removed,
and the pruned RBM was compared on the reconstruction of images with
that of the original RBM, as illustrated in Figure
\ref{fig:inv_prune_vis}. In Figure \ref{fig:prune_vis} we shows the
reconstruction of test images from RBMs in which low-scored features
were gradually removed.

\begin{figure}[ht]
  \centering
     \includegraphics[width=.8\textwidth]{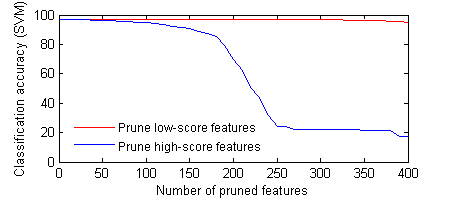}  
     \caption{Classification accuracy of a pruned RBM, starting with
       500 hidden units, on 10,000 MNIST test samples. The red line
       presents the pruning performance of low-scored features and the blue line
       presents the pruning performance of high-score features}
 \label{fig:c_test}
\end{figure}

Finally, in order to obtain accuracy measures, we have provided the
features obtained from the pruned RBMs as an input to an SVM
classifier. Figure \ref{fig:c_test} shows the drop in accuracy with
the gradual pruning of the RBM. In case of pruning low-scored
features, at first, some removal of units have produced a slight
increase in accuracy. Then, the results indicate that it is possible
to remove nodes and maintain performance almost unchanged (until
relevant units start to be removed, at which point accuracy
deteriorates, when more than half of the number of units is
removed). In case of pruning high-scored features, the accuracy
decreases significantly when more than $10\%$ of hidden units are
removed.
 
\subsection{Unsupervised Transfer Learning} 

In order to evaluate transfer learning in the context of images, we
have transferred sub-networks from an RBM trained on the
MNIST\footnote{\scriptsize \url{http://yann.lecun.com/exdb/mnist/}}
dataset to RBMs trained on the ICDAR\footnote{\scriptsize
  \url{http://algoval.essex.ac.uk:8080/icdar2005/index.jsp?page=ocr.html}}
and TiCC
datasets\footnote{\scriptsize\url{http://homepage.tudelft.nl/19j49/Datasets.html}}.

Below, we use MNIST$_{30k}$ and MNIST$_{05K}$ to denote datasets with
30,000 and 5,000 samples from the MNIST data collection. For the TiCC
collection, we denote TiCC$_d$ and TiCC$_a$ as the datasets of digits
and letters, respectively. TiCC$_c$ and TiCC$_w$ are character samples
from two different groups of writers. TiCC$_{w\_A}$ and TiCC$_{w\_B}$ are
from the same group but the latter has a much smaller training
set. Each column in Table \ref{tab:result_1} indicates a transfer
experiment, e.g. MNIST$_{30k}$ : ICDAR$_d$ uses the MNIST dataset as
source domain and ICDAR as target domain. The percentages show the
predictive accuracy on the target domain, as detailed in the sequel.

\begin{table}[ht]
\centering
\footnotesize{
\begin{tabular} {|l|c |c |c | c |}
\hline
&MNIST$_{30k}$ : ICDAR$_d$ & MNIST$_{30k}$ : TiCC$_{w\_A}$ & MNIST$_{05k}$ : TiCC$_a$ & MNIST$_{05k}$ : TiCC$_d$ 
\\
\hline
SVM                      & 39.04                       & 73.44                      & 59.16                       & 60.34\\
 
PCA STL                  &  39.38                      & 68.36                      & 57.90                       & 56.29\\

SC STL                   &  \textbf{46.23}             & 70.06                      & 55.82                            &  57.78\\

RBM STL                  &  30.47 $\pm$ 0.054                     & 72.88   $\pm$ 0.098                   & 58.13  $\pm $ 0.205                    & 62.08  $\pm$ 0.321\\
RBM                      & 37.63$\pm$ 0.505            & 75.20 $\pm$ 0.745          & \textbf{62.85 $\pm$ 0.079}  & 63.42 $\pm$ 0.090 \\
\hline
ASTL ($\alpha=0$)        & 36.66$\pm$ 0.495            & 76.49 $\pm$ 0.361          & \textbf{63.21 $\pm$ 0.134}  & 65.04 $\pm$ 0.330   \\
ASTL ($\alpha=1$)        &  40.43 $\pm$ 0.328 & \textbf{77.56 $\pm$ 0.564} & \textbf{63.00 $\pm$ 0.160}  &\textbf{65.82 $\pm$ 0.262}  \\
\hline
\end{tabular}
}
\caption{Transfer learning experimental results: each column indicates a 
transfer experiment, e.g. MNIST$_{30k}$ : ICDAR$_d$ uses the MNIST dataset as
 source domain and ICDAR as target domain. The percentages show
 the predictive accuracy on the target domain. Results for SVMs are provided
 as a baseline. For the "SVM" and "RBM" lines, there is no transfer; 
  for the other lines, transfer is carried out as described in Section 
\ref{sec:pre} and Section \ref{sec:learning}. The percentages show the average results with $95\%$ confidence interval}
\label{tab:result_1}
\end{table}

In Table \ref{tab:result_1}, we have measured classification
performance when using the knowledge from the source domain to guide
the learning in the target domain. For each domain, the data is
divided into training, validation and test sets. Each experiment is
repeated 50 times and we use the validation set to select the best
model (number of transferred sub-networks, number of added units,
learning rates and SVM hyper-parameters). The table reports the
average results on the test sets. The results show that the proposed
adaptive feature ranking and transfer learning method offers
statistically significant improvements on the training of RBMs in
three out of four cases tested, where the MNIST dataset is used as
source and either the ICDAR or TiCC dataset is used as target. As
expected, our method improves on the predictive accuracy of the
standard self-taught learning method for unsupervised transfer
learning.  In order to compare with transferring low-scored features,
we performed experiments follows what we have done with transferring
high-scored features, except that in this case the features are ranked
from low-scores to high-scores. We observed that the highest
accuracies were achieved only when a large number of high-scored
features are among those which have been transferred.

\begin{table}[ht]
\centering
\footnotesize{
\begin{tabular} {|l|c |c |c |}
\hline
& TiCC$_d$ : TiCC$_a$ & TICC$_a$ : TiCC$_d$ &  TICC$_c$ : TICC$_{w\_B}$ 
\\
\hline
SVM                      & 59.16                        & 60.34                        & 40.67   \\
RBM STL                   & 60.65 $\pm$ 0.075                        & 64.85  $\pm$ 0.227                       & 47.46 $\pm$ 0.260  \\
RBM                      & 62.85 $\pm$ 0.079            & 63.42 $\pm$ 0.090            & 44.28 $\pm$ 0.323  \\
\hline
ASTL ($\alpha =0$)        & 62.41 $\pm$ 0.166            & \textbf{66.10 $\pm$ 0.137}  & \textbf{51.07 $\pm$ 0.684}     \\

ASTL ($\alpha =1$)        & \textbf{63.16 $\pm$ 0.120}   &\textbf{66.25 $\pm$ 0.175}    & 43.10 $\pm$ 0.332 \\ 
\hline
\end{tabular}
}
\caption{Transfer learning experimental results for datasets in TiCC collection. 
 The percentages show the average  predictive accuracy on 
the target domain with $95\%$ confidence interval
}
\label{tab:results_2}
\end{table}

We also carried out experiments on the TiCC collection transferring
from characters to digits, digits to characters, and from group of
writers to group of other writers. The results in Table
\ref{tab:results_2} show that in two out of three experiments adaptive
learning and combining features did not gain advantages over
combination of selective features from source domain and features from
target domain. It may suggest that using selective combination of
features would be better and more efficient in a case that the source and
target domains are considerably similar(i.e TiCC$_c$:TICC$_{w\_B}$)
\begin{figure}[ht]
  \centering
    \subfigure[MNIST to ICDAR digits] {
      \includegraphics[width=0.3\textwidth]{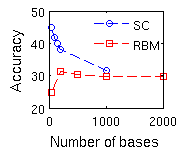}
    }
    \subfigure[MNIST to TiCC letters] {
      \includegraphics[width=0.3\textwidth]{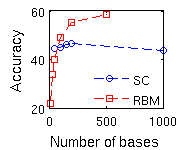}
    }
\\
    \subfigure[MNIST to TiCC digits]{
      \includegraphics[width=0.3\textwidth]{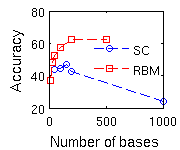}
    }
  \subfigure[MNIST to TiCC writers]{
      \includegraphics[width=0.3\textwidth]{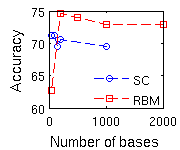}
   }

  \caption{Performance of self-taught learning using RBM and sparse-coder regarding to number of bases/hidden units}
\label{fig:selftaught}
\vskip -0.5cm
\end{figure}

To compare our transfer learning approach with self-taught learning
\cite{Raina_2007}, we train an RBM in the source domain and use it to
extract common features in the target domain for classification. In
the experiments where the domains have a close relation such as the
same type of data (digits in MNIST$_{05k}$:TiCC$_d$) or in the same
collections
(TiCC$_d$:TiCC$_a$,TiCC$_d$:TiCC$_d$,TiCC$_c$:TiCC$_{wB}$),
sefl-taught learning works very well especially when the training
dataset in the target domain is small (TiCC$_c$:TiCC$_{wB}$). We also
use sparse-coder \cite{Lee_2006} provided by
Lee\footnote{\url{http://ai.stanford.edu/~hllee/softwares/nips06-sparsecoding.htm}}
for self-taught learning as in \cite{Raina_2007}, except that instead
of using PCA for preprocessing data we apply the model directly to the
raw pixels since that is what has been done with the RBM. Figure
\ref{fig:selftaught} shows the performance of self-taught learning on
the datasets using RBMs and sparse-coder as feature learners.

\begin{figure}[ht]
  \centering
    \subfigure[MNIST to ICDAR] {
      \includegraphics[width=0.315\textwidth]{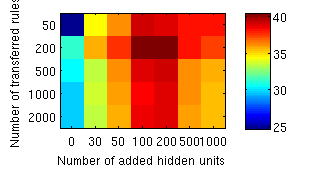}
    }
    \subfigure[MNIST to TiCC digits]{
      \includegraphics[width=0.315\textwidth]{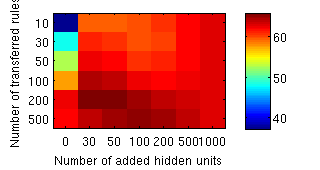}
    }
  \subfigure[MNIST to TiCC letters]{
      \includegraphics[width=0.315\textwidth]{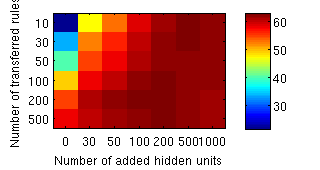}
   }
    \subfigure[MNIST to TiCC writers] {
      \includegraphics[width=0.315\textwidth]{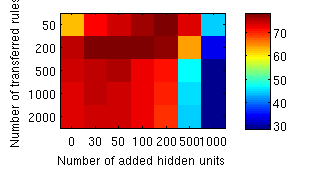}
    }
  \subfigure[TiCC letters to TiCC digits]{
      \includegraphics[width=0.315\textwidth]{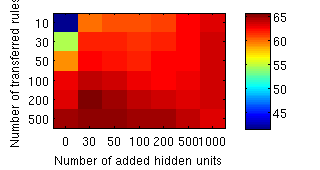}
   }
    \subfigure[TiCC digits to TiCC letters]{
      \includegraphics[width=0.315\textwidth]{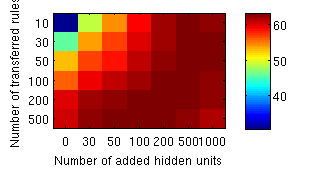}
    }
    \caption{Performance of learning with guidance for different
      numbers of transferred knowledge rules and additional
      hidden units. The colour-bars map accuracy to the colour of
      the cells as shown so that the hotter the color, the higher the accuracy.}
\label{fig:trans_add}
\vskip -1cm
\end{figure} 



With transfer, it is generally accepted that the performance
of the model in a target domain will depend on the quality of the
knowledge it received and the structure of the model. We then
evaluated performance of the model using different sizes of
transferred knowledge and number of units added to the hidden layer.
Figure \ref{fig:trans_add} shows that if the size of transferred
knowledge is too small, it will be dominated by the data from the
target domain. However, if the size of transferred knowledge is too
large it can cause a drop in performance since the model will try to
learn new knowledge mainly based on the transferred knowledge with
little knowledge from the target domain.

\section{Conclusion and Future Work}
We have presented a method and efficient algorithm for ranking and
selecting representations from a Restricted Boltzmann Machine trained
on a source domain to be transferred onto a target domain.  The method
is general in that the knowledge chosen by the ranking function does
not depend on its relation to any specific target domain, and it works
with unsupervised learning and knowledge-based transfer. Experiments
carried out using the MNIST, ICDAR and TiCC image datasets show that
the proposed adaptive feature ranking and transfer learning method
offers statistically significant improvements on the training of RBMs.

In this paper we focus on selecting features from shallow network
(RBM) for general transfer learning. In future work we are interested
in learning and transferring high-level features from deep networks
selectively for a specific domain.

\bibliographystyle{plain} 
\bibliography{bibio} 
\end{document}